\documentclass[10pt, a4paper]{article}

\usepackage[final]{lrec2026} 

\usepackage{amsmath}
\usepackage{amssymb}
\usepackage{booktabs}
\usepackage{multirow}
\usepackage{tikz}
\usepackage{graphicx}
\usepackage{float}
\usetikzlibrary{arrows.meta, positioning, fit, backgrounds, shapes.geometric, calc, decorations.pathreplacing}

\title{PAST-TIDE: Prototype-Anchored Statement Tuning\\
with Topic-Invariant Normalization for Stance Detection}

\name{Md.\ Shakhoyat Rahman Shujon$^{1}$, MD \ Jahid Hasan Jim$^{1}$, Md.\ Milon Islam$^{1}$,\\
\textbf{Md.\ Rezwanul Haque}$^{2}$, \textbf{Fakhri Karray}$^{2,3}$}
\address{$^{1}$Department of Computer Science and Engineering, Khulna University of Engineering \& Technology\\
        \texttt{skt104.shujon@gmail.com, mdjahidhasanjim277@gmail.com,}\\
        \texttt{milonislam@cse.kuet.ac.bd}
         \\
         $^{2}$Department of Electrical and Computer Engineering, University of Waterloo\\
         $^{3}$Department of Machine Learning, Mohamed bin Zayed University of Artificial Intelligence\\
         \texttt{\{rezwan, karray\}@uwaterloo.ca}}

\abstract{
We introduce \textbf{PAST-TIDE}, our stance detection system addressing both subtasks of the StanceNakba Shared Task at NakbaNLP@LREC-COLING 2026. The main idea is statement tuning. We redefine stance as cloze-style masked language modeling (MLM), letting a verbalizer map label words to stance categories through the pre-trained MLM head rather than appending a randomly initialized classification head. We complement this with prototypical contrastive learning, which uses learnable class prototypes for batch-size independent contrastive training, and topic-conditional layer normalization for cross-topic Arabic stance detection. \textbf{PAST-TIDE} achieves macro-F1 scores of 0.75 for Subtask~A and 0.74 for Subtask~B on the official leaderboard, indicating that minimal architectural additions to a pre-trained model can remain competitive in low-resource settings.
\\ \newline }

\newcommand\blfootnote[1]{%
  \begingroup
  \renewcommand\thefootnote{}
  \footnote{#1}%
  \addtocounter{footnote}{-1}
  \endgroup
}

\begin{document}


\maketitleabstract
\blfootnote{\textit{Proceedings of the 2nd International Workshop on Nakba Narratives as Language Resources (Nakba-NLP 2026) @ LREC 2026, pages 252--257 \\ 11 May 2026. \copyright ELRA Language Resources Association (ELRA), 2026}}
\enlargethispage{3pt}

\section{Introduction}
\label{sec:intro}

Understanding a person’s stance on a given topic remains a challenging problem in computational argumentation \citep{mohammad-etal-2016-semeval, kuccuk2020stance}. The field has moved from hand-crafted features to pre-trained language model fine-tuning \citep{ghosh-etal-2019-stance, li-caragea-2021-target} to prompt-based reformulations \citep{schick-schutze-2021-exploiting, gao-etal-2021-making}, yet multilingual stance detection remains underexplored. Most of the existing work focuses on benchmarks in English alone with thousands of labeled samples, leaving open the question of whether prompt-based methods remain effective when labels are limited and the target language is not English. The eNakba Shared Task \citep{stance_nakba_2026} addresses exactly the above gap. It defines two subtasks: Subtask~A, stance at the actor-level in English (Pro-Palestine/Pro-Israel/Neutral), and Subtask~B, stance cross-topic in Arabic (Favor/Against/ None). Both tasks use relatively small datasets, with 1,401 and 1,205 labeled samples, split into 70/15/15 for train, development, and test. A standard [CLS] classifier head lacks any pre-trained initialization of its own, and that pre-training/fine-tuning gap degrades performance at this scale.

Our architecture, \textbf{PAST-TIDE}, is based on one idea: the classistance classification capacity already exists withinined MLM head, and the task is to access it rather than replace it. The major contributions of this paper are as follows. First, statement tuning redefines stance as cloze-style MLM, bypassing the randomly initialized [CLS] head with a multi-token verbalizer that averages predictions across register-diverse label words. Second, prototypical contrastive learning (PCL) replaces in-batch negatives with $K{=}3$ class prototypes, where these prototypes are updated by gradients and remain persistent, giving stable anchors even at small batch sizes. Third, topic-conditional layer normalization (T-CLN), inspired by conditional normalization in vision, generates per-topic normalization parameters for cross-topic transfer.

On the official leaderboard, \textbf{PAST-TIDE} placed $11^{th}$ in Subtask~A (macro-F1: 0.75) and $7^{th}$ in Subtask~B (macro-F1: 0.74), scoring macro-F1 of 0.79 on both during the testing phase. It is observed that the architectures architectures with more parameters (up to 2.6M additional weights) scored 30\%--43\% lower during development, whereas our proposed model uses fewer than 1.3M trainable parameters and no the classification path. The source code is publicly available.\footnote{\url{https://github.com/Shakhoyat/PAST-TIDE}}

The paper is organized as follows. In section 2, we review related work, and in section 3, we explain our proposed architecture. Sections 4 and 5  demonstrate experiments and results, and Section 6 provides the conclusion.

\section{Related Work}
\label{sec:related}

Stance detection has been studied primarily in English, from the SemEval-2016 shared task \citep{mohammad-etal-2016-semeval} to cross-target generalization \citep{allaway-mckeown-2020-zero}. \citet{kuccuk2020stance} provide a survey of this field up to 2020. Arabic stance work is still limited, with the MARASTA corpus \citep{charfi2024marasta} as one of the few multi-dialectal resources. Prompt-based classification via cloze-style MLM \citep{schick-schutze-2021-exploiting, gao-etal-2021-making} shows consistent few-shot gains, but its application to stance detection, particularly in Arabic, has gained little attention. Contrastive learning for text classification typically relies on in-batch negatives \citep{khosla2020supervised}, which becomes inconsistent at small batch sizes. Prototypical networks \citep{snell2017prototypical} offer an alternative, though the episodic formulation does not suit standard fine-tuning directly. Multilingual encoders like mDeBERTa-v3 \citep{he2021debertav3} have made cross-lingual transfer practical, yet combining them with prompt-based methods under data scarcity remains challenging. \textbf{PAST-TIDE} is designed for this problem: statement tuning for stance in a low-resource multilingual setting, with persistent prototypes replacing in-batch contrastive signals, and lightweight topic conditioning for Arabic cross-topic transfer.

\section{System Description}
\label{sec:method}

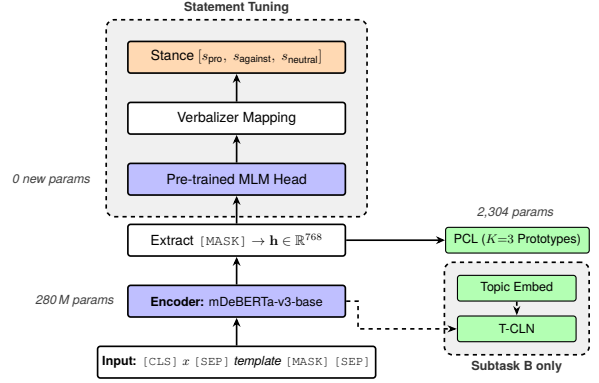
\begin{figure}[t]
\centering
\resizebox{\columnwidth}{!}{%
\begin{tikzpicture}[
    node distance=0.7cm,
    box/.style={draw, very thick, rounded corners=2pt, minimum height=0.8cm,
                minimum width=5.5cm, align=center, font=\small\sffamily,
                fill=#1},
    box/.default=white,
    sbox/.style={draw, thick, rounded corners=2pt, minimum height=0.65cm,
                 minimum width=3.0cm, align=center, font=\footnotesize\sffamily,
                 fill=#1},
    sbox/.default=white,
    grp/.style={draw, very thick, dashed, rounded corners=6pt, inner sep=9pt},
    arr/.style={-{Stealth[length=2.3mm, width=2mm]}, very thick},
    darr/.style={-{Stealth[length=2.4mm, width=1.8mm]}, very thick, dashed},
    lbl/.style={font=\footnotesize\sffamily\itshape, text=black!75},
]


\node[box] (input)
    {\textbf{Input:} \texttt{[CLS]}~$x$~\texttt{[SEP]}~\textit{template}~\texttt{[MASK]}~\texttt{[SEP]}};

\node[box=blue!25] (enc) [above=of input]
    {\textbf{Encoder:} mDeBERTa-v3-base};

\node[box, font=\normalsize\sffamily] (mask) [above=of enc]
    {Extract \texttt{[MASK]} $\rightarrow$ $\mathbf{h} \in \mathbb{R}^{768}$};

\node[box=blue!25, font=\normalsize\sffamily] (mlm) [above=of mask]
    {Pre-trained MLM Head};

\node[box, font=\normalsize\sffamily] (verb) [above=of mlm]
    {Verbalizer Mapping};

\node[box=orange!30, font=\normalsize\sffamily] (stance) [above=of verb]
    {Stance~$[s_{\text{pro}},\; s_{\text{against}},\; s_{\text{neutral}}]$};

\draw[arr] (input) -- (enc);
\draw[arr] (enc)   -- (mask);
\draw[arr] (mask)  -- (mlm);
\draw[arr] (mlm)   -- (verb);
\draw[arr] (verb)  -- (stance);

\node[sbox=green!30, right=2.5cm of mask, minimum width=3.633cm] (pcl) {PCL ($K{=}3$ Prototypes)};
\draw[arr] (mask.east) -- (pcl.west);

\node[sbox=green!30, right=2.8cm of enc, yshift=-0.7cm] (tcln) {T-CLN};
\node[sbox=green!30, above=0.35cm of tcln] (temb) {Topic Embed};
\draw[darr] (temb) -- (tcln);
\draw[darr] (enc.east) -- ++(0.3,0) |- (tcln.west);  
\node[lbl, anchor=east] at ([xshift=-0.2cm]enc.west)   {280\,M params};
\node[lbl, anchor=east] at ([xshift=-0.8cm]mlm.west)   {0 new params};
\node[lbl, anchor=south west] at ([yshift=0.08cm, xshift=0.7cm]pcl.north west) {2{,}304 params};

\begin{scope}[on background layer]
    \node[grp, fill=gray!12, inner sep=15pt,
          fit=(mlm)(verb)(stance),
          label={[font=\small\sffamily\bfseries, text=black!80]above:Statement Tuning}] (stmt) {};

    \node[grp, fill=gray!12,
          fit=(tcln)(temb),
          label={[font=\small\sffamily\bfseries, text=black!80]below:Subtask B only}] {};
\end{scope}

\end{tikzpicture}%
}
\caption{\textbf{PAST-TIDE} architecture. The cloze-wrapped input is passed to the encoder and the \texttt{[MASK]} hidden state passes through the pre-trained MLM head and verbalizer to generate stance logits. PCL prototypes branch from the same hidden state (auxiliary loss). T-CLN (Subtask~B only) modifies the full encoder output before extraction.}
\label{fig:architecture}
\end{figure}

The proposed architecture covers both subtasks, with T-CLN switching on only for Subtask~B (Fig.~\ref{fig:architecture}). The idea is that a pre-trained MLM head already understands words like \textit{support} and \textit{against}, and stance labels are just vocabulary. Hence, we use the head directly for classification rather than replacing it. The methods, including PCL, T-CLN, and R-Drop then solve specific failure modes: gradient collapse in small batches, cross-topic drift, and a weak decision boundary.

\subsection{Backbone}

Our encoder is mDeBERTa-v3-base \citep{he2021debertav3} (280M parameters, hidden dimension 768), chosen because it includes a pre-trained MLM head that we reuse directly. A single checkpoint is used for both subtasks.

\subsection{Statement Tuning}
\label{sec:statement-tuning}

Stance detection is redesigned as a cloze-style MLM task, instead of adding a randomly initialized classification head to the [CLS] token. For each input text $x$ and target $t$, we construct a statement:

\vspace{0.15cm}
\noindent\fbox{\parbox{0.93\columnwidth}{\small
\texttt{[CLS]} $x$ \texttt{[SEP]} Regarding $t$, this author's stance is \texttt{[MASK]}. \texttt{[SEP]}
}}
\vspace{0.15cm}

\noindent The phrasing ``this author's stance is [MASK]'' was chosen because the MLM head has learnt many sentences of the form ``X is [ADJ]'' during pre-training, so the slot naturally provides adjective-like predictions where SEP stands for Separator Token. Other variations, we tried placing \textit{pro} or \textit{against} in syntactic positions where they rarely occur in natural text, and this caused lower performance.

At the \texttt{[MASK]} position, the pre-trained MLM head generates vocabulary logits and a \textit{verbalizer} maps these logits to class probabilities by aggregating log-probabilities of semantically relevant tokens as mentioned in (\ref{eq:verbalizer}).
\begin{equation}
\label{eq:verbalizer}
s_k = \frac{1}{|V_k|} \sum_{w \in V_k} \log P_{\text{MLM}}(w \mid \texttt{[MASK]})
\end{equation}

\noindent where, $V_k$ is the verbalizer set for class $k$, $V_0$ = \{\textit{support, favor, pro, yes}\} (Pro/Favor), $V_1$ = \{\textit{oppose, against, anti, no}\} (Against), $V_2$ = \{\textit{neutral, unclear, none}\} (Neutral). Multi-subword tokens use the first subword (primary semantics in SentencePiece). This setup requires \textit{zero new parameters}. The MLM head is already pre-trained and semantically aligned with the verbalizer tokens, which is exactly what makes it functioning at the given sample scale.

\textbf{Verbalizer Design:} Each class gets 3--4 tokens spanning different registers: \textit{support} and \textit{favor} are formal, \textit{pro} and \textit{yes} are informal. Mixing registers helps because social media text varies widely in formality. We tried Arabic tokens for Subtask~B, but mDeBERTa has a limited Arabic vocabulary, and the performance decreased. English tokens, when passed through the shared embedding space, worked well for both languages.

\subsection{Prototypical Contrastive Learning}
\label{sec:pcl}

We introduce PCL \cite{khosla2020supervised} to structure the representation space, where $K{=}3$ learnable class prototypes $\{\mathbf{p}_1, \mathbf{p}_2, \mathbf{p}_3\} \in \mathbb{R}^{768}$. The loss pushes each sample's \texttt{[MASK]} embedding toward its ground-truth prototype as shown in (\ref{eq:pcl}).
\begin{equation}
\label{eq:pcl}
\mathcal{L}_{\text{PCL}} = -\log \frac{\exp(\text{sim}(\mathbf{h}, \mathbf{p}_y) / \tau)}{\sum_{j=1}^{K} \exp(\text{sim}(\mathbf{h}, \mathbf{p}_j) / \tau)}
\end{equation}

\noindent where $\tau{=}0.1$ and $\mathbf{h}$ is the L2-normalized \texttt{[MASK]} hidden state. Unlike Supervised Contrastive Learning (SupCon), whose gradient quality depends on in-batch negatives (only ${\sim}5.3$ expected at batch size 8), PCL's denominator always has exactly $K{=}3$ terms, keeping optimization consistent even at batch size 1. In practice, this turned out to be more than we initially expected. Early experiments with SupCon showed fold-level F1 varying by up to $\pm 8$ points, which we found was due to random batches where one class was severely underrepresented. PCL almost entirely eliminated this variance.

\subsection{Topic-Conditional Layer Normalization}
\label{sec:tcln}

Subtask~B requires stance detection across two Arabic topics with distinct vocabulary distributions. The first-order transformed static parameters $\gamma, \beta$ of standard layer normalization are replaced by T-CLN \citep{de2017modulating, huang2017arbitrary} with topic-conditioned parameters as in (\ref{eq:tcln-embed}) and (\ref{eq:tcln}).
\begin{align}
\mathbf{e}_t &= \text{TopicEmbed}(t_{\text{id}}) \in \mathbb{R}^{64} \label{eq:tcln-embed} \\
\text{T-CLN}(x, t) &= \text{MLP}_\gamma(\mathbf{e}_t) \odot \frac{x - \mu}{\sigma} + \text{MLP}_\beta(\mathbf{e}_t) \label{eq:tcln}
\end{align}

\noindent where, $\text{MLP}_\gamma, \text{MLP}_\beta$ are two-layer networks ($64{\to}768{\to}768$, Tanh), initialized as identity ($\gamma_t{\approx}1, \beta_t{\approx}0$). T-CLN is applied after the last transformer layer and before the MLM head; it is only activated for Subtask~B. The identity initialization is important: it means T-CLN starts as a no-operation and gradually learns topic-specific shifts during training. We found that the normalization parameters diverged early and destabilized the MLM head without this proper initialization.

\subsection{R-Drop Regularization and Training}

R-Drop \citep{liang2021rdrop} performs two forward passes with different dropout masks and minimizes their symmetric KL divergence $\mathcal{L}_{\text{R-Drop}}$, regularizing the decision boundary in low-resource settings. The total training loss is calculated in (\ref{eq:total-loss}).

\begin{equation}
\label{eq:total-loss}
\mathcal{L} = \mathcal{L}_{\text{focal}} + 0.1 \cdot \mathcal{L}_{\text{PCL}} + 1.0 \cdot \mathcal{L}_{\text{R-Drop}}
\end{equation}

\noindent where, $\mathcal{L}_{\text{focal}}$ is a class-weighted focal loss \citep{lin2017focal} ($\gamma{=}2.0$, label smoothing 0.1) applied to the verbalizer logits, averaged over both R-Drop passes. Training proceeds in two stages: Stage~1 (epochs 0--1) freezes the encoder and MLM head, updating only PCL prototypes and T-CLN. Stage~2 (epochs 2--9) unfreezes everything with layer-wise learning rate decay (\citealp{clark-etal-2020-electra}) (decay 0.95/layer, base LR $2{\times}10^{-5}$, $5\times$ multiplier for heads).

\section{Experimental Setup}
\label{sec:setup}

\textbf{Dataset:} The StanceNakba 2026 dataset \citep{stance_nakba_2026,charfi2024marasta} provides 1,401 English samples for Subtask~A and 1,205 Arabic samples for Subtask~B, each split into 70/15/15 for train, development, and test sets. Subtask~A covers actor-level stance toward the Israeli-Palestinian conflict (Pro-Palestine/Pro-Israel/Neutral). The training data contains 980 samples (327/326/327 per class) with 210 development and 211 test samples. Subtask~B is organized around two topics, \textit{normalization with Israel} (577 samples) and \textit{refugees in Jordan} (628 samples), with labels Favor/Against/Neither. The training split has 843 samples (286/298/259 per class), with 181 development and 181 test cases. The class distributions are almost balanced, though Subtask~B's \textit{Neither} class is somewhat smaller, making it difficult to learn. Both datasets are collected from social media, hence posts range from single-sentence to multi-paragraph arguments, and code-switching appears frequently in the Arabic data.

\textbf{Augmentation and Ensembling:} We increase the training set by double using back-translation with MarianMT \citep{junczys-dowmunt-etal-2018-marian}, pivoting through German\footnote{German's divergent word order (verb-final subordinate clauses, free constituent order) forces the round-trip to paraphrase rather than copy, unlike Romance pivots whose syntactic similarity to English yields near-identical back-translations. The Helsinki-NLP opus-machine translation EN$\leftrightarrow$DE are also among the highest quality available models.} for Subtask~A and English for Subtask~B. Evaluation uses 5-fold stratified cross-validation. We save the best checkpoint per fold using early stopping with a patience of 3, then average the class probabilities across folds during test phase. Arabic inputs go through a preprocessing step before tokenization: Alef and Ya forms are unified, diacritics are removed, and Tatweel characters are deleted.

\textbf{Implementation:} We utilized Dual NVIDIA T4 GPUs (2$\times$16\,GB). The batch size is set to $8{\times}4{=}32$ after gradient accumulation, with a maximum sequence length of 256 tokens, FP16, and gradient checkpointing to stay within memory limits. We optimize with AdamW, increasing linearly for first 10\% of steps and then decaying on a cosine schedule.


\section{Results and Analysis}
\label{sec:results}

Table~\ref{tab:official} reports official leaderboard scores. All rows are server-evaluated submissions. 

\begin{table}[t]
\centering
\small
\setlength{\tabcolsep}{4pt}
\begin{tabular}{lcrrrr}
\toprule
\textbf{Subtask} & \textbf{Rank} & \textbf{F1} & \textbf{Acc} & \textbf{P} & \textbf{R} \\
\midrule
A (Actor-Level)  & \#11 & 0.745 & 0.749 & 0.763 & 0.749 \\
B (Cross-Topic)  & \#7  & 0.741 & 0.740 & 0.745 & 0.743 \\
\bottomrule
\end{tabular}
\caption{Official leaderboard results (organizer evaluation). F1\,=\,macro-F1, Acc\,=\,Accuracy, P\,=\,macro-precision, and R\,=\,macro-recall.}
\label{tab:official}
\end{table}


\begin{table}[t]
\centering
\small
\begin{tabular}{lccc}
\toprule
\textbf{Methods} & \textbf{A} & \textbf{B} & \textbf{Mean} \\
\midrule
Dual-Encoder ({[CLS]})     & 0.50 & 0.64 & 0.57 \\
Target-Calibrated ({[CLS]})    & 0.66 & 0.70 & 0.68 \\
Disentangled (MLM)  & 0.46 & 0.64 & 0.55 \\
Frame-Anchored (SSA)   & 0.48 & 0.42 & 0.45 \\
w/o T-CLN (MLM)     & 0.75 & 0.74 & 0.75 \\
\hline
\textbf{PAST-TIDE (MLM)} & \textbf{0.79} & \textbf{0.79} & \textbf{0.79} \\
\midrule
\end{tabular}
\caption{Macro-F1 scores of testing-phase submissions. SSA\,=\,Soft Semantic Anchors. }
\label{tab:main-results}
\end{table}


Table~\ref{tab:main-results} shows the development results of the proposed architecture. During the testing phase, \textbf{PAST-TIDE} scored 0.79 macro-F1 on both subtasks, 16\% above our best [CLS]-based system. The official leaderboard settled lower (0.75 / 0.74), consistent with the expected generalization gap. The testing phase scores were computed on a subset of the development data, while the final evaluation used the held-out test partition. The highest achievement came from statement tuning itself: the MLM head already has representations for words like \textit{support} and \textit{against}. Replacing SupCon with PCL reduced fold-level F1 variance by up to 8 points. The architectures with more parameters scored worse: the Disentangled variant added 2.6M parameters and reduced performance by 30\%, and freezing the MLM head decreased performance by 43\%.

\begin{table}[t]
\centering\small
\setlength{\tabcolsep}{3.5pt}
\begin{tabular}{l ccc ccc}
\toprule
& \multicolumn{3}{c}{\textbf{Subtask A}} & \multicolumn{3}{c}{\textbf{Subtask B}} \\
\cmidrule(lr){2-4} \cmidrule(lr){5-7}
& \textcolor{green}{Pro-P} & \textcolor{green}{Pro-I} & \textcolor{green}{Neut} & \textcolor{blue}{Fav} & \textcolor{blue}{Agn} & \textcolor{blue}{Nei} \\
\midrule
\textcolor{green} {Pro-P}\,/\,\textcolor{blue}{Fav} & \textbf{56} & 8 & 6 & \textbf{46} & 13 & 3 \\
\textcolor{green} {Pro-I}\,/\,\textcolor{blue}{Agn} & 7 & \textbf{60} & 3 & 7 & \textbf{50} & 6 \\
\textcolor{green} {Neut}\,/\,\textcolor{blue}{Nei}  & 11 & 3 & \textbf{56} & 3 & 12 & \textbf{41} \\
\bottomrule
\end{tabular}
\caption{Dev-set confusion matrices (210\,/\,181 samples). Rows: actual, columns: predicted, Pro-P\,=\,Pro-Palestine, Pro-I\,=\,Pro-Israel, Neut\,=\,Neutral, Fav\,=\,Favor, Agn\,=\,Against, and Nei\,=\,Neither.}
\label{tab:confusion}
\end{table}

\textbf{Confusion Matrix:} Table~\ref{tab:confusion} demonstrates the confusion matrices for both subtasks. In Subtask~A, most errors come from Neutral posts being misclassified as Pro-Palestine (11 of 14 total Neutral errors), while only 3 are predicted as Pro-Israel. This happens because words about ceasefire and civilian deaths are similar to words in Pro-Palestinian posts, so the model cannot separate them properly. The architecture confuses the two main classes almost equally (8 vs. 7). This means it can separate them, but has trouble with neutral posts. In Subtask~B, the model over-predicts \textit{against}: 13 pro instances are misclassified as \textit{against} compared to 7 in the opposite direction, and 12 neutral instances are also predicted as \textit{against}. We believe this happens because strong pro-stance Arabic writing uses emphatic words, dialect forms, and strong emotion, which looks similar to opposition. Hence, the model gets confused because both sides use similar words, more than in English.

\textbf{Cross-Topic Transfer:} It is observed that removing T-CLN has no effect on Subtask~A (single English topic), but drops Subtask~B by 6\% in the testing phase. This shows that the two Arabic topics are different in the representation space, and conditional normalization bridges them.

\textbf{Parameter Overhead:} In terms of overhead, Subtask~A adds just 2,304 new parameters (the PCL prototypes); Subtask~B adds ${\sim}1.28$M for T-CLN. The MLM head and verbalizer contribute zero new weights, which is arguably the most important design choice. The entire classification capacity comes from weights that were already trained on hundreds of gigabytes of multilingual text.

\section{Conclusion}
\label{sec:conclusion}

\textbf{PAST-TIDE} obtained 0.75 and 0.74 macro-F1 on the official Subtask~A and~B leaderboards (macro-F1: 0.79 on both during the testing phase), by reusing rather than replacing the pre-trained MLM head. Our simpler proposed models outperformed larger ones with added parameters. The task design is more important than adding more modules to the model at this small data scale. Each module of the proposed architecture works naturally: statement tuning to any label set with real vocabulary words, PCL to any setting where batch sizes are too small for in-batch negatives, and T-CLN to any multi-domain problem. In the future, we plan to automate verbalizer selection via gradient-based token search and test on benchmarks with finer granularity of the labels.

\section*{Bibliographical References}
\vspace{-20pt}
\label{sec:reference}

\bibliographystyle{lrec2026-natbib}
\bibliography{references}

\appendix
\section{Hyperparameters}
\label{sec:appendix-hyperparams}

\begin{table}[H]
\centering
\small
\begin{tabular}{ll}
\toprule
\textbf{Hyperparameter} & \textbf{Value} \\
\midrule
Backbone & mDeBERTa-v3-base \\
Max sequence length & 256 \\
Effective batch size & $8 \times 4 = 32$ \\
Learning rate (backbone) & $2 \times 10^{-5}$ \\
Head LR multiplier & $5\times$ \\
LLRD decay per layer & 0.95 \\
Weight decay & 0.01 \\
Warmup ratio & 0.1 \\
Focal loss $\gamma$ & 2.0 \\
Label smoothing & 0.1 \\
PCL $\tau$ / weight & 0.1 / 0.1 \\
R-Drop weight & 1.0 \\
T-CLN topic dim & 64 (Subtask B) \\
Training stages & 2 frozen + 8 fine-tuned \\
K-fold / patience & 5 / 3 \\
\bottomrule
\end{tabular}
\caption{Hyperparameter configurations for \textbf{PAST-TIDE}.}
\label{tab:hyperparams}
\end{table}

\section{Limitations}
\label{sec:limitations}

Our design choices are based on cross-iteration failure analysis rather than a controlled ablation within a single architecture variant, owing to compute limits (dual T4, 9h on Kaggle). This means the improvements from each component may be mixed with other changes between runs, and a proper ablation study would make the results stronger. Verbalizer tokens were chosen by intuition about stance-relevant vocabulary. an automated search (e.g., gradient-based token selection) might find better tokens, particularly for Arabic where our lexical intuitions are weaker. Finally, the shared task data covers two specific geopolitical topics, and we have not yet tested whether the architecture transfers well to domains with different label semantics or number of classes.



\section{Ethics Statement}
\label{sec:ethics}

\textbf{PAST-TIDE} infers stance from surface-level linguistic cues; the predictions carry no implication about the authors' own views on the topics involved. All data were used strictly according to the shared task guidelines.

\section{Acknowledgements}
\label{sec:acknowledgements}

We are grateful to the StanceNakba 2026 organizers for providing the datasets and evaluation infrastructure. All experiments were run on Kaggle's free GPU tier. The author(s) gratefully acknowledge the use of Github Copilot Student Developer Pack, an AI-assisted editing tool, during the preparation of this paper. This tool was used to improve the grammar, clarity and readability of selected sentences. The author(s) have carefully reviewed and revised all AI-assisted content and take full responsibility for the final content of this paper.

\end{document}